%% file: iclr2027_conference.tex
\documentclass{article} 
\usepackage{iclr2027_conference,times}
\usepackage{graphicx}
\usepackage{array}
\usepackage{booktabs}
\usepackage{tabularx}
\usepackage{caption}
\usepackage{multirow}

\input{math_commands.tex}

\usepackage{hyperref}
\usepackage{xurl}

\title{VoxelTTO: Voxel-Aligned Feed-Forward 3D Gaussian Splatting with Test-Time Optimization}

\author{Yibin Zhao\thanks{These authors contributed equally.} \\
East China University of Science and Technology \\
Shanghai, China \\
\texttt{Y20230063@mail.ecust.edu.cn} \\
\And
Yihan Pan\footnotemark[1] \\
Shanghai Open University \\
Shanghai, China \\
\texttt{email@ou.sh.cn} \\
\AND
Wenli Yang, Jun Nan \& Jianjun Yi\thanks{Corresponding author.} \\
East China University of Science and Technology \\
Shanghai, China \\
\texttt{\{Y30260694,Y20240039\}@mail.ecust.edu.cn, jjyi@ecust.edu.cn}
}

\iclrfinalcopy 
\begin{document}

\maketitle

\begin{figure*}[!h]
    \centering
    \includegraphics[width=\linewidth]{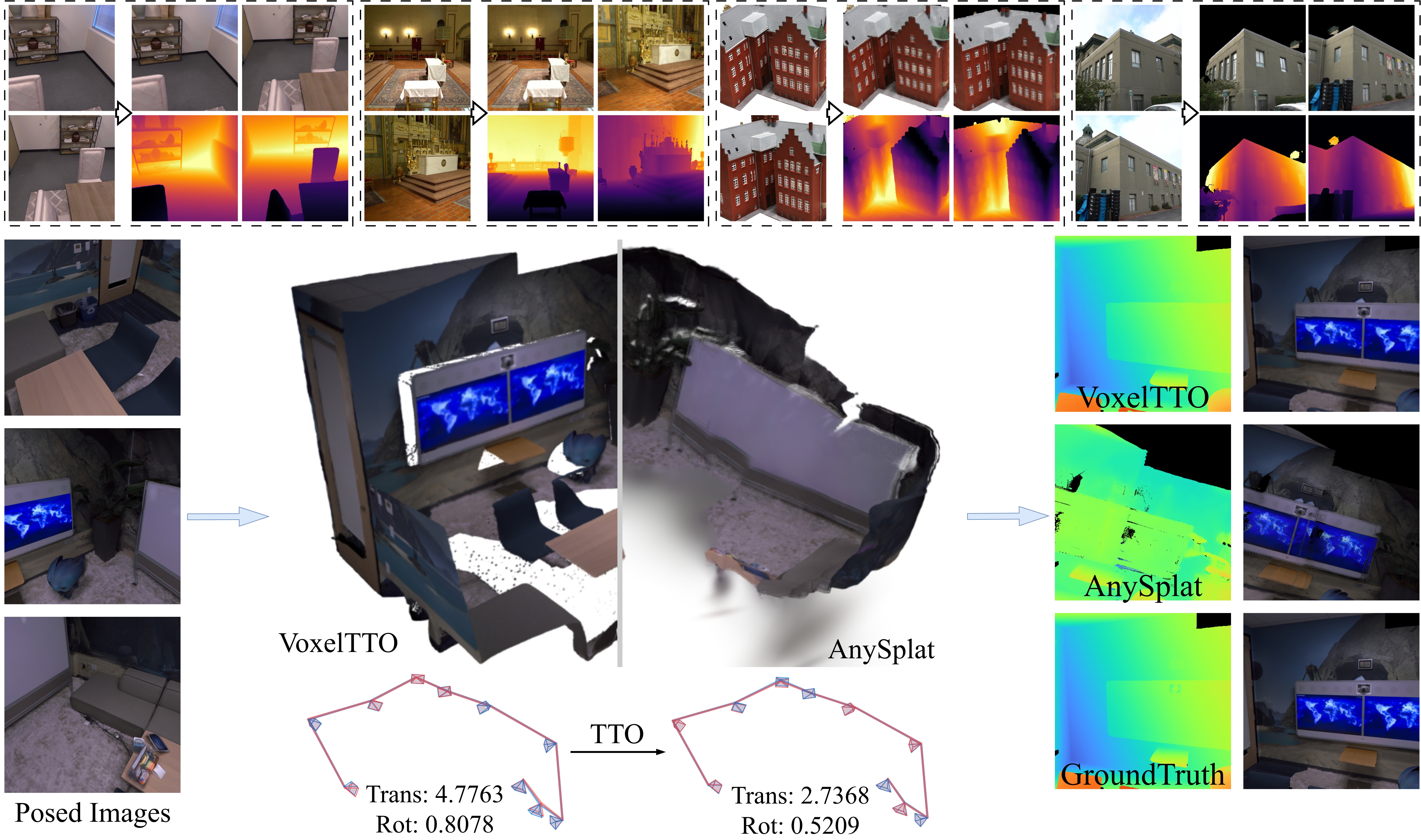}
    \caption{VoxelTTO directly regresses a geometrically accurate Gaussian-splatting scene from posed images. It precisely aligns camera poses and improves RGB and depth rendering in both input and test views across indoor--outdoor and synthetic--realworld scenes.}\label{toutu}
    \label{fig:toutu}
\end{figure*}

\begin{abstract}
Recent feed-forward 3D Gaussian Splatting (3DGS) methods typically regress pixel-aligned Gaussian primitives, often causing excessive overlap and artifacts, while inaccuracies in predicted camera poses can lead to misalignment in novel-view synthesis (NVS).
We present VoxelTTO, a feed-forward framework for reconstructing geometrically accurate 3DGS scenes from an arbitrary number of images and optional camera parameters. VoxelTTO aggregates dense image features into a global voxel representation and decodes Gaussians from voxel features, breaking the pixel-to-Gaussian correspondence. To exploit known camera parameters while keeping the pretrained visual foundation model (VFM) parameters frozen, we introduce test-time optimization (TTO) that adapts lightweight LoRA modules using pose supervision. We further replace vanilla 3DGS rasterization with stochastic solid volume rendering during training and inference, improving geometric fidelity. Training updates only the voxel-aligned Gaussian reconstruction modules, requiring 80 GPU hours. Experiments on Replica, Tanks and Temples, and DTU demonstrate improved RGB-D NVS and camera-pose estimation relative to prior methods.
\end{abstract}

\section{Introduction}

Conventional 3D reconstruction systems from images rely on multistage pipelines comprising feature matching \cite{sarlin2020superglue,detone2018superpoint} and structure from motion (SfM) \cite{colmapsfm}. Recent VFMs \cite{wang2025vggt,dust3r_cvpr24} offer an end-to-end alternative, using a large Transformer to directly predict 3D attributes from arbitrary image collections.

Despite their strong geometric priors, point-map and depth-map representations produced by VFMs often fail to capture fine details and photorealistic appearance. By contrast, 3DGS \cite{kerbl3DGS}, and their extensions \cite{Huang2DGS2024,scaffoldgs} use richer scene representations to deliver accurate NVS. However, these methods commonly recover camera parameters with SfM before performing costly per-scene optimization. Some approaches directly predict 3DGS scenes \cite{xu2024depthsplat,charatan23pixelsplat,chen2024mvsplat}, but they typically require accurate camera parameters and can degrade in complex scenes. More recent methods \cite{jiang2025anysplat,depthanything3,ye2025yonosplatneedmodelfeedforward} extend the feed-forward VFMs to 3DGS. They regress pixel-aligned 3DGS scenes directly from unposed images, but their geometric accuracy remains limited. Several VFMs encode camera parameters within the backbone to exploit additional priors and improve prediction accuracy \cite{keetha2026mapanything,depthanything3,jiang2025anysplat}. Nevertheless, the resulting accuracy remains insufficient, and persists substantial misalignment during test-view synthesis.

In this work, we present VoxelTTO, a voxel-aligned feed-forward 3DGS framework with test-time optimization (TTO). VoxelTTO reconstructs geometrically accurate 3DGS scenes from arbitrary numbers of images and optional camera parameters. When camera priors are available, it supports accurate NVS from test views. To reduce excessive overlap and artifacts, VoxelTTO departs from conventional pixel-aligned feed-forward 3DGS, upsamples patch tokens from a frozen VFM backbone with DPT and fuses them into global voxels. A sparse convolutional U-Net encodes the voxel features, and an appearance-geometry decoupled self-splitting MLP regresses Gaussian primitive parameters. We replace 3DGS rasterization with stochastic solid volume rendering, which provides stronger geometric supervision during optimization and improves geometric fidelity for NVS .

To exploit camera priors, TTO inserts LoRA modules into the VFM backbone. Pose-supervised adaptation of these modules improves alignment between predicted and ground-truth camera parameters. During training, we freeze the VFM backbone and optimize only the Voxel-Aligned Feed-Forward 3DGS module. This design introduces 75.3M trainable parameters, significantly reduces training cost to 80 GPU hours, and is compatible with any VGGT-style VFMs.

Our main contributions are as follows:
\begin{itemize}
    \item We propose VoxelTTO, a voxel-aligned feed-forward 3DGS reconstruction framework that recovers geometrically accurate scenes from arbitrary image collections and optional camera parameters.
    \item We introduce voxel-aligned feed-forward 3DGS and stochastic solid volume rendering supervision to reduce excessive overlap and artifacts while strengthening geometry.
    \item We introduce pose-supervised TTO to exploit camera priors and improve alignment.
    \item Extensive experiments across multiple datasets demonstrate improved camera-pose estimation and more accurate RGB-D rendering in both input and test views compared to existing feed-forward 3DGS methods.
\end{itemize}

\section{Related Works}
 \subsection{Optimization-Based Novel-View Synthesis}

Neural radiance fields (NeRF) \cite{mildenhall2020nerf} model scene density and view-dependent color with neural fields, but ray-wise volume rendering is computationally expensive. 3DGS \cite{kerbl3DGS} represents scenes with anisotropic Gaussian primitives and enables real-time NVS; subsequent work improves its geometric representation \cite{Huang2DGS2024,Dai2024GaussianSurfels} and optimization \cite{huang2025fatesgs,guedon2023sugar}. Stochastic solid volume rendering is equivalent to 3DGS rasterization under suitable conditions \cite{2023Objectsasvolumes,zhang2026gggs,GaussianWrapping} and improves depth and surface reconstruction.

These optimization-based methods require accurate camera parameter, which are typically estimated with SfM pipelines\cite{colmapsfm,colmapmvs,wang2024vggsfm,dust3r_cvpr24}. 
Several methods jointly optimize camera parameters and 3DGS scenes, but generally require sequential images \cite{keetha2024splatam,cf3dgs} or depth maps \cite{gsicpslam}. 
Vanilla 3DGS initializes primitives by KNN distances from a colored point cloud and optimizes each scene independently \cite{kerbl3DGS}. Subsequent research
s directly regress a 3DGS scene from images and camera parameters without optimization \cite{charatan23pixelsplat,chen2024mvsplat,xu2024depthsplat}.

\subsection{Visual Foundation Models}
Image-based reconstruction has traditionally relied on separate SfM and dense-reconstruction stages. Recent VFMs directly regress 3D scene information from multiple views. Dust3R \cite{dust3r_cvpr24} uses self and cross-attention to infer aligned point maps and camera parameters from image pairs.
VGGT alternates single image attention with global attention to regress depth maps, point maps, and camera parameters, from arbitrary views. Later work increases model accuracy\cite{depthanything3,wang2026vggtomega} and improves order invariance \cite{wang2025pi3}.
Several VFMs \cite{jiang2025anysplat,depthanything3,ye2025yonosplatneedmodelfeedforward} further support direct regression of 3DGS. Their Gaussian primitives are usually pixel-aligned with depth maps, which inevitably introduces excessive overlap and rendering artifacts. Directly regressing Gaussians from fixed tokens \cite{tokengs2026,li2026querysplat} alleviates this issue but commonly sacrifices geometric accuracy.

VFM geometry is predicted across views and may not strictly satisfy multiview constraints. Test-time optimization (TTO) has shown benefits for Vision Transformer tasks \cite{han2025vittt} and been applied to VFMs \cite{elflein2026vggttt}. Recent TCO inserts LoRA modules into attention layers and constrains them using known camera parameters and depth maps, improving pose estimation and dense reconstruction \cite{tco}. Unlike approaches that encode depth maps or camera parameters into the backbone \cite{keetha2026mapanything,depthanything3,jiang2025anysplat}, TCO adapts lightweight modules without changing the base VFM architecture.
\vspace{-0mm}
\section{Method}
\vspace{-4mm}
\begin{figure}[!h]
    \centering
    \includegraphics[width=\linewidth]{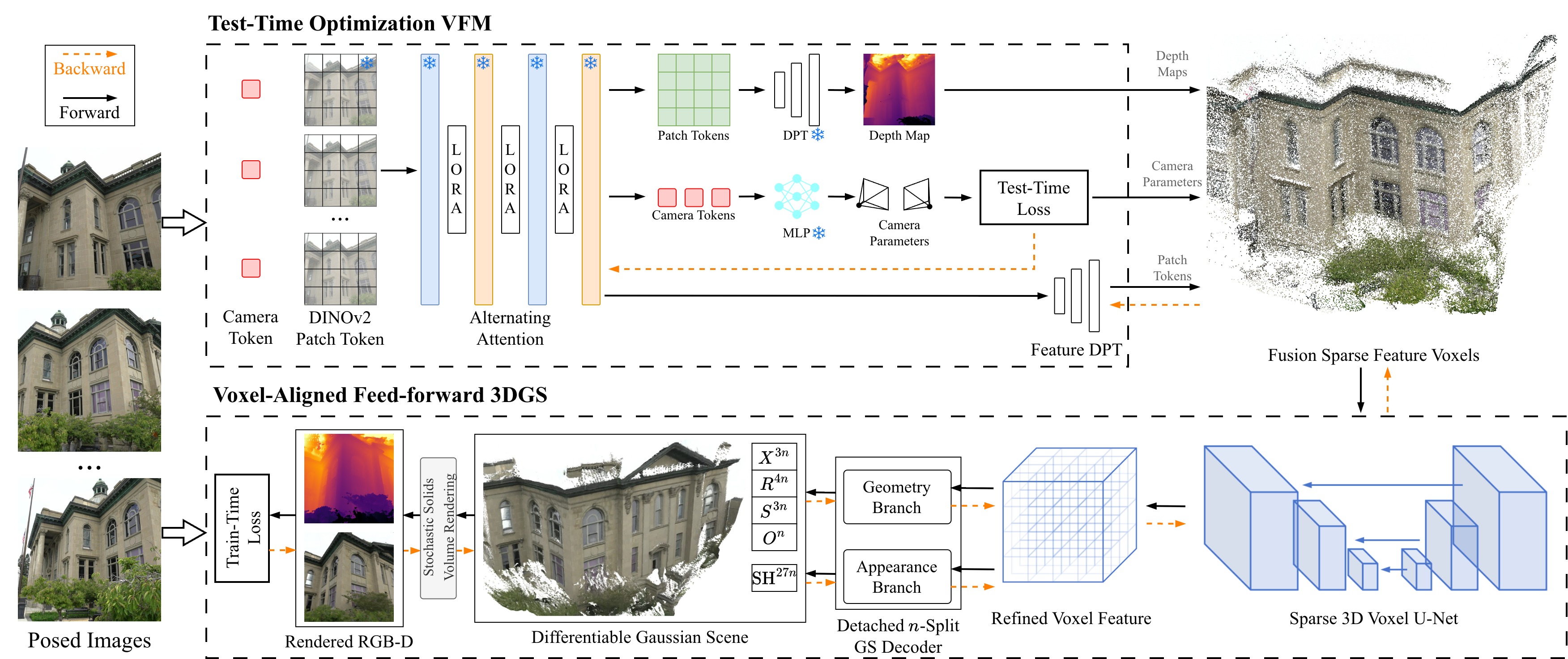}
    \caption{Overview of the VoxelTTO pipeline.}
    \vspace{-4mm}
    \label{fig:pipeline}
\end{figure}
We propose VoxelTTO, a feed-forward method that predicts a 3DGS scene from multiview images and optional camera parameters. By aggregating dense features into voxels and processing them with a sparse-convolutional U-Net, VoxelTTO delivers high geometric fidelity and appearance performance. TTO further addresses the misalignment commonly observed when VGGT-style VFMs predict Gaussian-splatting scenes, improving accuracy in test views.

\subsection{Preliminaries and Challenges}

VFMs \cite{wang2025vggt,dust3r_cvpr24} use a large Transformer $\Phi_{\theta}$ to learn cross-view visual representations from $M$ uncalibrated images $\mathbf{I} = \left\{\mathbf{I}_{m}\right\}_{m=1}^{M}$. Separate prediction heads $\Gamma_{\psi}$ recover 3D scene information, including point maps $\mathbf{P}_{m}\in\mathbb{R}^{H\times W\times3}$, depth maps $\mathbf{D}_{m}\in\mathbb{R}^{H\times W}$, and camera parameters $\mathbf{g}_{m}$, as shown in \autoref{eq:vfm_formulation}.
\begin{equation}
\left(\mathbf{P}_{m},\mathbf{D}_{m},\mathbf{g}_{m}\right)=\Gamma_{\psi}\left(\Phi_{\theta}\left(\mathbf{I}\right)\right), \qquad m=1,\ldots,M
\label{eq:vfm_formulation}
\end{equation}



3DGS VFMs typically upsample patch tokens with DPT to obtain pixel-aligned Gaussian attributes (center $\mu_{q} $, opacity $\sigma_{q}$, rotation quaternion $r_{q}$, scale $s_{q}$, and color $c_{q}$ ). Such methods often lack multiview consistency and create \textbf{excessive overlap, producing artifacts}. 
Direct token-to-Gaussian methods commonly have lower geometric accuracy \cite{li2026querysplat}.

To exploit camera priors, several VFMs \cite{keetha2026mapanything,depthanything3,jiang2025anysplat} encode ground-truth camera parameters $\mathbf{g}_{\text{gt}}$ as tokens before the backbone and incorporate them during inference. Although fusing auxiliary camera parameters with RGB features improves performance, it changes the architecture and prevents direct reuse of unconstrained VFMs. Moreover, the inferred camera parameters $\mathbf{g}_{m}$ can still differ substantially from $\mathbf{g}_{\text{gt}}$. Even after Umeyama SIM(3) alignment \cite{umeyamaalign}, this error causes substantial \textbf{caemra misalignment} in subsequent NVS.

\subsection{Voxel-Aligned Feed-Forward 3DGS}
\subsubsection{Network Architecture}
VoxelTTO uses DINOv2 \cite{oquab2023dinov2} to extract multiview patch tokens and employs alternating attention to model the scene. We add LoRA to the alternating-attention layers for TTO, as detailed in \autoref{TTO_section}.
The frozen VFM backbone first predicts a depth map $\mathbf{D}_{m}$, camera parameters $\mathbf{g}_{m}$, and confidence map $\mathbf{K}_{m}$ for each view. We add a feature DPT that upsamples patch tokens into dense pixel features $\mathbf{F}_{m}\in\mathbb{R}^{H\times W\times C}$. We then back-project pixels into world coordinates, obtaining a dense feature point cloud $\mathcal{P}=\{(\mathbf{p}_l,\mathbf{a}_l,\mathbf{f}_l,\kappa_l)\}_{l=1}^{L}$, where $\mathbf{p}_l$, $\mathbf{a}_l$, $\mathbf{f}_l$, and $\kappa_l$ denote point coordinate, color, feature, and confidence, respectively.
To reduce memory use and avoid excessive overlap, we aggregate the point cloud into voxel features. Let $\mathcal{P}_v$ denote the points within voxel $v$. Its feature center is $\mathbf{x}_v={\sum_{l\in\mathcal{P}_v}\kappa_l\mathbf{p}_l}/{\sum_{l\in\mathcal{P}_v}\kappa_l}$, which is not constrained to the regular voxel center and thus preserves geometric accuracy. Voxel color $\mathbf{a}_v$ and feature $\mathbf{f}_v$ are also confidence-weighted averages. We define voxel confidence as $\kappa_v=\min_{l\in\mathcal{P}_v}\kappa_l$ to reduce the impact of unreliable points. A 3D U-Net refines the voxel features for scene understanding. Sparse convolutions \cite{spconv2022} avoid computation in empty voxels.

To increase the representational capacity of each voxel, we decode every voxel feature $\mathbf{f}_v$ into $n$ Gaussian primitives, allowing one voxel to represent complex local detail. To avoid entangling geometry and appearance during optimization \cite{li2026querysplat}, we use separate geometry and appearance branches, $\mathcal{D}_{\mathrm{geo}}$ and $\mathcal{D}_{\mathrm{app}}$.
Centers, rotation quaternions, scales, and opacities of $n$ Gaussian primitives are decoded by $\mathcal{D}_{\mathrm{geo}}$. 
Second-order spherical-harmonic coefficients are decoded by $\mathcal{D}_{\mathrm{app}}$.
\subsubsection{Stochastic Solid Volume Rendering}
To further improve the geometric accuracy of NVS, we use stochastic solid volume rendering\cite{2023Objectsasvolumes} instead of vanilla 3DGS rasterization. With an appropriate vacancy function $v$, volume rendering of the stochastic solid is equivalent to 3DGS rasterization \cite{zhang2026gggs}, as shown in \autoref{eqn_volrender}, where $\boldsymbol{\omega}$ is the ray direction, $\sigma(\mathbf{x},\boldsymbol{\omega})$ is the stochastic solid attenuation coefficient at point $\mathbf{x}$ along the ray, $T(t)$ is accumulated transmittance at position $t$, and $\mathbf{c}$ is the directional color.
\begin{equation}
    \begin{aligned}
        \label{eqn_volrender}
        \sigma(\mathbf{x}, \boldsymbol{\omega})= \left| \boldsymbol{\omega}^{\top} \nabla \log v(\mathbf{x}) \right|  &= \frac{\left| \boldsymbol{\omega}^{\top} \nabla v(\mathbf{x}) \right|}{v(\mathbf{x})}\qquad
        T(t) = \exp\left(-\int_{t_n}^{t}\sigma(\mathbf{x}(s), \boldsymbol{\omega})\, ds\right)\\
        \mathbf{C} &= \int_{t_n}^{t_f} T(t)\,\sigma(\mathbf{x}(t), \boldsymbol{\omega})\,\mathbf{c}(\mathbf{x}(t), \boldsymbol{\omega})\, dt
    \end{aligned}
\end{equation}

An appropriate vacancy $v$ is required to obtain images equivalent to 3DGS rasterization. GGGS\cite{zhang2026gggs} proves that the renderings are equivalent when $v(\mathbf{x}) = \sqrt{1 - G(\mathbf{x})}$. The accumulated transmittance is then computed as shown in \autoref{eqn_leijitoushelv}, where $t_i^{*}$ is the location of the Gaussian maximum along the ray.
\begin{equation}
    \begin{aligned}
        \label{eqn_leijitoushelv}
        T_i(t) = \begin{cases} \sqrt{1 - G_i(t)}, & t \leq t_i^{*}, \\ \dfrac{1 - G_i(t_i^{*})}{\sqrt{1 - G_i(t)}}, & t > t_i^{*}. \end{cases}
    \end{aligned}
\end{equation}

For depth rendering, we use the median depth, defined by the point where $T(t_\text{med}) = 0.5$ with a fixed number of bisection iterations. During backpropagation, we differentiate the fixed condition $T(t_\text{med}) = 0.5$ with respect to Gaussian parameters $\theta$, as shown in \autoref{eqn_volrender_back}.
\begin{equation}
    \begin{aligned}
        \label{eqn_volrender_back}
        \frac{\partial T}{\partial t}\frac{\partial t_{\mathrm{med}}}{\partial \theta} + \frac{\partial T}{\partial \theta} = 0
    \end{aligned}
\end{equation}

\vspace{-3mm}

\begin{figure}[!h]
    \centering
    \includegraphics[width=0.9\linewidth]{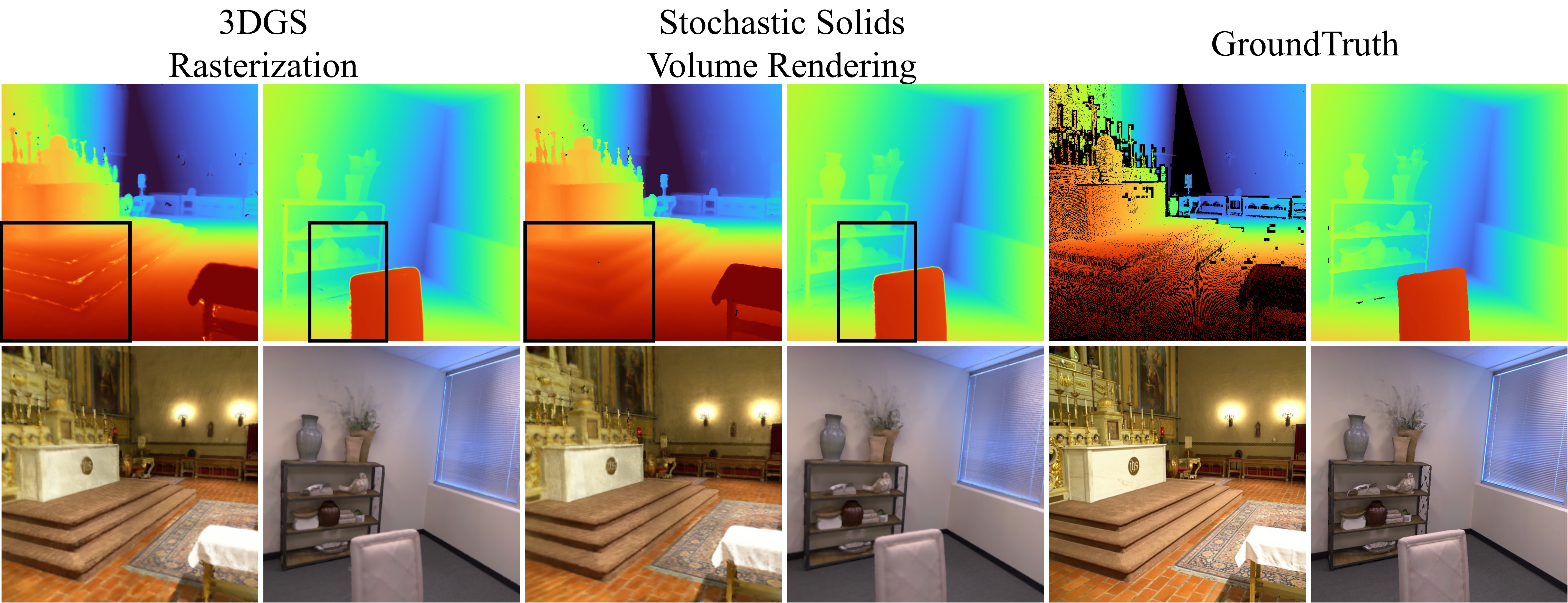}
    \caption{Geometric advantages of stochastic solid volume rendering over 3DGS rasterization.}
    \label{fig:volrender_okvis}
\end{figure}
As shown in \autoref{fig:volrender_okvis}, stochastic solid volume rendering produces RGB images that are nearly identical to 3DGS rasterization for the same Gaussian scene. For depth rendering, however, it offers a clear geometric advantage by reducing holes and blurred boundaries.

\subsection{Test-Time Optimization for VFMs}\label{TTO_section}
\begin{figure}[!h]
    \centering
    \includegraphics[width=\linewidth]{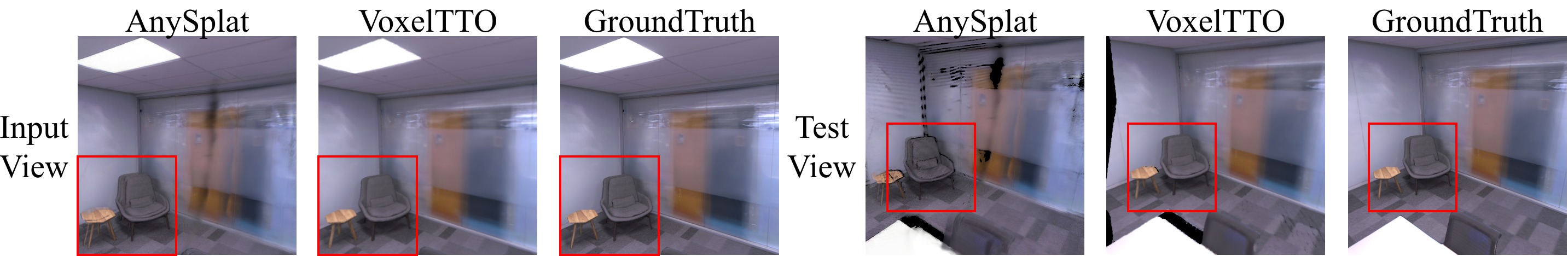}
    \caption{Test-view misalignment after injecting camera parameters.}
    \vspace{-4mm}
    \label{fig:vis_align}
\end{figure}

For reconstruction from posed images, some VFMs encode camera parameters with an MLP and inject them directly into the Transformer. Although this approach improves geometric accuracy, the improvement is insufficient for test views NVS. As shown in \autoref{fig:vis_align}, AnySplat maintains reasonable input view quality despite artifacts, but produces misalignment in test views.

To address this issue, we introduce LoRA \cite{hu2021lora} into the VFM backbone and perform TTO. Following the DA3 architecture \cite{depthanything3}, our network has 40 attention layers. Layers 1--12 use single-view attention, whereas layers 13--40 alternate between single-view and global attention. We add LoRA to the $\mathbf{Q}$, $\mathbf{K}$, and $\mathbf{V}$ projections in layers 13--40. Each layer shares a down-projection matrix $\mathbf{A}$, while their up-projection matrices $\mathbf{B}$ are independent.

When aligned camera parameters are provided, we compute the optional TTO loss after camera decoding and update the LoRA parameters by backpropagation. After fixed iterations, the backbone produces more accurate camera parameters aligned with groundtruth cameras.
VGGT-style VFMs usually use the first image as the scene coordinate frame, and image-based reconstruction has scale ambiguity. We therefore align poses before calculating the loss. We fix the first ground-truth camera pose to the identity matrix. The rotation loss is constructed from the relative rotation between predicted rotations $\mathbf{R}^{p}_{i}$ and ground-truth rotations $\mathbf{R}^{g}_{i}$, as shown in \autoref{eq:rotation_loss}. 
\begin{equation}
\mathcal{L}_{\mathrm{rot}}=\frac{1}{2M}\sum_{i=1}^{M}\left[3-\operatorname{tr}\!\left(\left(\mathbf{R}^{p}_{i}\right)^{\top}\mathbf{R}^{g}_{i}\right)\right].
\label{eq:rotation_loss}
\end{equation}
\vspace{-2mm}

To address scale ambiguity, we normalize each translation set by mean scale $s=M^{-1}\sum_{i=1}^{M}\lVert\mathbf{t}_i\rVert_2$, computed separately for the predicted and ground-truth translations. The scale-normalized translation loss is then calculated by SmoothL1.
We also supervise camera intrinsics. The VFM parameterizes intrinsics as $k=(f_x,f_y,c_x,c_y)$, and we directly apply SmoothL1 to the predicted and ground-truth values. The camera-based TTO objective is
\(L_{\mathrm{TTO}} =
\lambda_{\mathrm{pose}}
\left(
L_{\mathrm{rot}}
+w_tL_{\mathrm{trans}}
\right)
+
\lambda_K L_{\mathrm{intrinsics}}\).

\subsection{Training}
\subsubsection{Datasets}
We train VoxelTTO on ScanNet \cite{dai2017scannet}, Infinigen \cite{infinigen2023infinite}, ARKitScenes \cite{baruch2021arkitscenes}, and DL3DV-10K \cite{ling2024dl3dv}. These datasets span indoor and outdoor environments as well as synthetic and real-world scenes. 
All datasets contain sequential frames. For each dataset, we define a minimum and maximum frame gap, randomly select the interval between the first and last frame within that range, and sample frames from it.

\subsubsection{Training Details}
Our architecture follows DA3-GIANT, uses $L=40$ alternating-attention layers as described in \autoref{TTO_section}, and is initialized with pretrained weights \cite{depthanything3}. The full model has 1.39B parameters, of only 75.3M parameters are trainable
We optimize for 30K iterations with Adam, an initial lr of $3\times10^{-6}$, cosine annealing, 2K warm-up iterations, and a minimum lr of $10^{-8}$. Each sample contains 3--16 frames resized to $224\times448$. Training takes 40 hours on 2 RTX PRO 6000. 

As the VFM remains frozen, only the voxel-aligned feed-forward 3DGS module is optimized. The loss contains rendered-depth loss $\mathcal{L}_{\mathrm{depth}}$ and rendered-RGB loss $\mathcal{L}_{\mathrm{rgb}}$, as shown in \autoref{eq:training_loss}.
\begin{equation}
\begin{aligned}
\mathcal{L}
= \mathcal{L}_{\mathrm{depth}} + \mathcal{L}_{\mathrm{rgb}}
= \lambda_{\mathrm{\ell^d_1}}\mathcal{L}_{\mathrm{depth}}^{\mathrm{\ell_1}}
+ \lambda_{\mathrm{align}}\mathcal{L}_{\mathrm{depth}}^{\mathrm{align}}
 + \lambda_{\ell_1}\mathcal{L}_{\mathrm{rgb}}^{\ell_1}
+ \lambda_{\mathrm{ssim}}\mathcal{L}_{\mathrm{rgb}}^{\mathrm{ssim}}
\end{aligned}
\label{eq:training_loss}
\end{equation}
Here, $\mathcal{L}_{\mathrm{depth}}^{\mathrm{\ell_1}}$ supervises rendered depth with ground-truth depth, whereas $\mathcal{L}_{\mathrm{depth}}^{\mathrm{align}}$ enforces alignment with VFM-predicted depth. $\mathcal{L}_{\mathrm{rgb}}^{\ell_1}$ and $\mathcal{L}_{\mathrm{rgb}}^{\mathrm{ssim}}$ are the $\ell_1$ and SSIM \cite{wang2004ssim} losses between rendered and ground-truth images. We set $\lambda_{\mathrm{\ell^d_1}}$, $\lambda_{\mathrm{align}}$, $\lambda_{\ell_1}$, and $\lambda_{\mathrm{ssim}}$ to 0.25, 0.25, 0.9, and 1.0.

\section{Experiments}
\subsection{Baselines}
We compare VoxelTTO with state-of-the-art (SOTA) baselines from two families: VGGT-style VFMs \cite{wang2025vggt} and pixelSplat-style \cite{charatan23pixelsplat} feed-forward 3DGS. VFM baselines include AnySplat \cite{jiang2025anysplat}, YoNoSplat \cite{ye2025yonosplatneedmodelfeedforward}, DepthAnything3 \cite{depthanything3}, and MapAnything\cite{keetha2026mapanything}. All support reconstruction from unposed images. AnySplat and DepthAnything3 support camera-parameter encoding.
The initialization of vanilla 3DGS is used on MapAnything.
\cite{kerbl3DGS}. pixelSplat-style methods typically require camera parameters; we evaluate VolSplat \cite{wang2025volsplat} and MVSplat \cite{chen2024mvsplat}.
Experiments are conducted on the Replica~\cite{replica19arxiv}, Tanks and Temples~\cite{tatdataset}, and DTU~\cite{dtu} datasets, including indoor and outdoor, synthetic and real-world, scene-level and object-level scenarios.

\subsection{RGB Novel-View Synthesis}
\label{nvs_exp}
\begin{figure*}[!h]
    \centering
    \includegraphics[width=\linewidth]{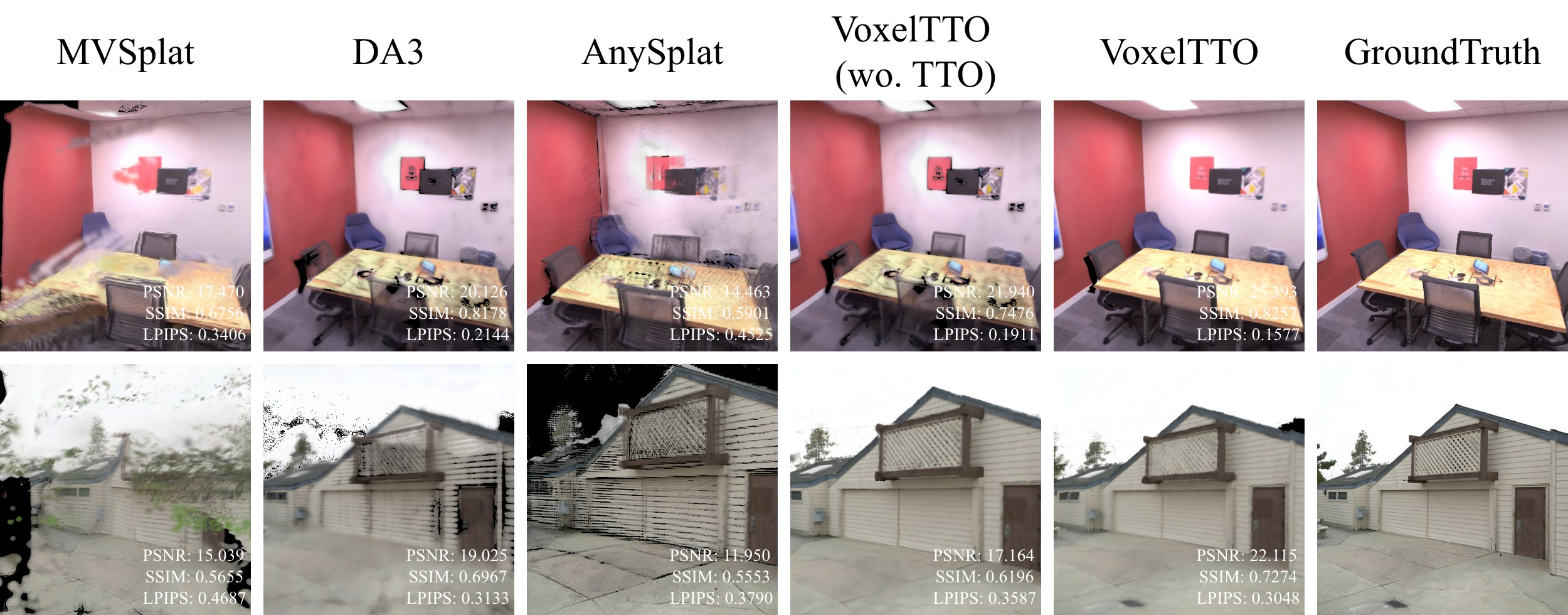}
    \caption{Qualitative RGB novel-view synthesis results for test views.}
    \vspace{-2.5mm}
    \label{fig:vis_nvs}
\end{figure*}

Following prior works\cite{kerbl3DGS}, we compare RGB NVS performance across VoxelTTO and the baselines. We use PSNR (P), SSIM (S) \cite{wang2004ssim}, and LPIPS (L) \cite{lpips}.
As shown in \autoref{tab:nvs_quantitative} and \autoref{fig:vis_nvs}, VoxelTTO performs strongly in both input and test views. TTO reduces pose-estimation error from the VFM backbone and substantially improves test-view NVS.
\vspace{-2mm}

\begin{table*}[!h]
    \centering
    \scriptsize
    \caption{Quantitative RGB novel-view synthesis results for input and test views.}
    \label{tab:nvs_quantitative}
    \renewcommand{\arraystretch}{0.8}
    \setlength{\tabcolsep}{1pt}
    \newcommand{\nvsTableScale}{1.25}
    \makebox[\textwidth][c]{\scalebox{\nvsTableScale}{%
    \begin{tabularx}{0.8\textwidth}{l*{12}{>{\centering\arraybackslash}X}}
        \toprule
        \textbf{Method} & \multicolumn{6}{c}{\textit{Replica Dataset}} & \multicolumn{6}{c}{\textit{TAT Dataset}} \\
        \cmidrule(lr){2-7} \cmidrule(lr){8-13}
        & \multicolumn{3}{c}{Input Views} & \multicolumn{3}{c}{Test Views} & \multicolumn{3}{c}{Input Views} & \multicolumn{3}{c}{Test Views} \\
        \cmidrule(lr){2-4} \cmidrule(lr){5-7} \cmidrule(lr){8-10} \cmidrule(lr){11-13}
        & P$\uparrow$ & S$\uparrow$ & L$\downarrow$ & P$\uparrow$ & S$\uparrow$ & L$\downarrow$ & P$\uparrow$ & S$\uparrow$ & L$\downarrow$ & P$\uparrow$ & S$\uparrow$ & L$\downarrow$ \\
        \midrule
        VolSplat       & 16.15 & 0.667 & 0.471 & 15.44 & 0.636 & 0.467 & 15.45 & 0.595 & 0.419 & 16.33 & \textbf{0.733} & \textbf{0.241} \\
        MVSplat        & 19.97 & 0.759 & 0.318 & 19.41 & 0.736 & 0.319 & 16.48 & 0.615 & 0.380 & 15.05 & 0.522 & 0.460 \\
        AnySplat       & 23.25 & 0.825 & 0.200 & 17.78 & 0.681 & 0.359 & 17.34 & 0.645 & \underline{0.311} & 12.71 & 0.440 & 0.478 \\
        YoNoSplat      & 24.52 & 0.809 & 0.262 & 11.57 & 0.635 & 0.432 & 13.61 & 0.477 & 0.516 & 13.29 & \underline{0.660} & \underline{0.304} \\
        DepthAnything3 & 23.64 & 0.837 & 0.208 & 22.82 & 0.817 & 0.243 & 19.20 & 0.621 & 0.348 & \underline{17.68} & 0.556 & 0.413 \\
        MapAnything    & 14.35 & 0.579 & 0.533 & 17.95 & 0.757 & 0.227 & 12.66 & 0.497 & 0.570 & 14.66 & 0.592 & 0.381 \\
        Ours (w/o TTO) & \underline{29.28} & \underline{0.890} & \underline{0.174} & \underline{25.83} & \underline{0.827} & \underline{0.196} & \textbf{23.74} & \textbf{0.767} & \textbf{0.247} & 14.71 & 0.485 & 0.475 \\
        Ours           & \textbf{29.91} & \textbf{0.896} & \textbf{0.170} & \textbf{28.54} & \textbf{0.876} & \textbf{0.172} & \underline{20.89} & \underline{0.656} & 0.347 & \textbf{19.23} & 0.565 & 0.391 \\
        \bottomrule
    \end{tabularx}
    }}%
\end{table*}

\vspace{-2mm}
\subsection{Depth Novel-View Synthesis}
\vspace{-4mm}
\begin{figure*}[!h]
    \centering
    \includegraphics[width=0.99\linewidth]{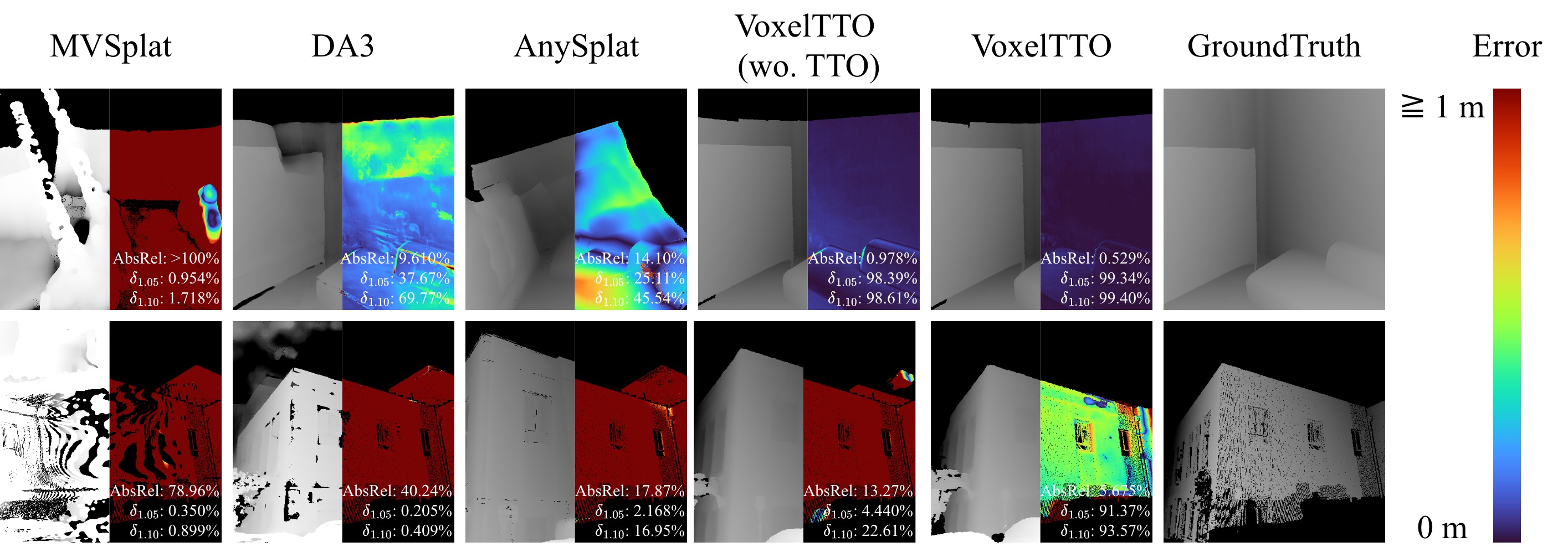}
    \caption{Qualitative depth novel-view synthesis results for test views.}
    \label{fig:vis_depth}
\end{figure*}
To assess the geometric accuracy, we compare depth NVS performance with each baseline. We report the absolute error $\mathrm{RMSE}$ (cm), relative error $\mathrm{Rel}$, and depth accuracy $\delta_{\tau}$ at several thresholds. 
Relative error (AbsRel) is adapted for relative depth error: $\mathrm{Rel}=(1/N)\sum_{i\in\mathcal{P}}(|\hat{d}_i-d_i|/d_i)$.
Depth accuracy $\delta_{\tau}$ is the proportion of pixels satisfying $\max(\hat{d}_i/d_i, d_i/\hat{d}_i)<\tau$.
As shown in \autoref{fig:vis_depth} and \autoref{tab:depth_nvs_quantitative}, VoxelTTO's geometric accuracy is higher than the baselines in both input and test views. 
TTO mitigates the misalignment caused by VFM and substantially improves test-view geometry.

\vspace{-1mm}
\begin{table*}[!h]
   \footnotesize
    \centering
    \caption{Quantitative depth novel-view synthesis results for input and test views.}
    \label{tab:depth_nvs_quantitative}
    \renewcommand{\arraystretch}{0.7}
    \setlength{\tabcolsep}{3pt}
    \begin{tabularx}{0.9\textwidth}{l*{10}{>{\centering\arraybackslash}X}}
        \toprule
        \textbf{Method} & \multicolumn{5}{c}{Input Views} & \multicolumn{5}{c}{Test Views} \\
        \cmidrule(lr){2-6} \cmidrule(lr){7-11}
        & RMSE$\downarrow$ & Rel$\downarrow$ & $\delta_{1.05}$$\uparrow$ & $\delta_{1.10}$$\uparrow$ & $\delta_{1.25}$$\uparrow$ & RMSE$\downarrow$ & Rel$\downarrow$ & $\delta_{1.05}$$\uparrow$ & $\delta_{1.10}$$\uparrow$ & $\delta_{1.25}$$\uparrow$ \\
        \midrule
        \multicolumn{11}{l}{\textit{Replica Dataset}} \\
        VolSplat       & 102.8 & 30.42 & 11.88 & 22.55 & 46.78 & 202.2 & 63.89 & 0.812 & 2.443 & 6.859 \\
        MVSplat        & 216.0 & 69.26 & 1.984 & 3.894 & 9.261 & 1649 & 446.1 & 11.84 & 25.77 & 42.37 \\
        AnySplat       & 33.72 & 9.582 & 59.05 & 76.88 & 86.14 & 35.92 & 10.78 & 45.43 & 67.09 & 89.86 \\
        YoNoSplat      & 23.47 & 7.281 & 51.07 & 75.58 & 94.78 & 242.1 & 78.73 & 0.304 & 0.692 & 2.643 \\
        DepthAnything3 & 125.5 & 43.27 & 28.30 & 46.40 & 57.85 & 172.0 & 38.69 & 64.18 & 71.19 & 76.29 \\
        MapAnything    & 13.30 & 3.559 & 82.69 & 93.74 & 98.12 & 37.96 & 10.15 & 44.80 & 68.37 & 86.92 \\
        Ours (w/o TTO) & \underline{7.394} & \underline{0.777} & \underline{98.57} & \underline{99.22} & \underline{99.66} & \underline{9.032} & \underline{1.115} & \underline{98.14} & \underline{98.83} & \underline{99.38} \\
        Ours           & \textbf{6.167} & \textbf{0.688} & \textbf{98.66} & \textbf{99.29} & \textbf{99.72} & \textbf{6.945} & \textbf{0.866} & \textbf{98.86} & \textbf{99.24} & \textbf{99.58} \\
        \midrule
        \multicolumn{11}{l}{\textit{TAT Dataset}} \\
        VolSplat       & 490.7 & 33.73 & 9.984 & 19.13 & 41.25 & 69.15 & 31.05 & 9.058 & 18.01 & 42.70 \\
        MVSplat        & 894.3 & 71.98 & 0.911 & 1.809 & 4.739 & 143.7 & 76.57 & 0.343 & 0.699 & 2.203 \\
        AnySplat       & 117.4 & 4.793 & 70.01 & 90.98 & 96.95 & 499.1 & 360.3 & 5.405 & 15.37 & 30.40 \\
        YoNoSplat      & 511.5 & 37.06 & 8.018 & 15.69 & 33.54 & 83.57 & 47.59 & 7.480 & 15.38 & 38.70 \\
        DepthAnything3 & 456.9 & 41.44 & 18.76 & 36.71 & 60.34 & 299.4 & 30.49 & 3.934 & 8.181 & 23.29 \\
        MapAnything    & 139.9 & 6.560 & 67.01 & 81.11 & 94.77 & 512.4 & 399.1 & 15.28 & 28.06 & 31.31 \\
        Ours (w/o TTO) & \textbf{82.07} & \textbf{2.079} & \textbf{96.35} & \textbf{97.45} & \textbf{98.49} & \underline{63.93} & 7.035 & \underline{52.39} & \underline{78.84} & \underline{95.44} \\
        Ours           & \underline{90.63} & \underline{2.875} & \underline{90.12} & \underline{95.83} & \underline{98.15} & \textbf{46.22} & \textbf{3.235} & \textbf{86.75} & \textbf{95.16} & \textbf{98.33} \\
        \bottomrule
    \end{tabularx}
\end{table*}
\vspace{-2mm}

\subsection{Pose Estimation}
\vspace{-2mm}
\begin{table}[!h]
    \centering
    \newcommand{\poseTableScale}{1.2}
    \newcommand{\surfaceTableScale}{1.2}
    \begin{minipage}[t]{0.65\textwidth}
        \centering
        \scriptsize
        \captionof{table}{Quantitative camera-pose estimation results.}
        \label{tab:pose_quantitative}
        \renewcommand{\arraystretch}{0.8}
        \setlength{\tabcolsep}{1.5pt}
        \scalebox{\poseTableScale}{%
        \begin{tabularx}{0.85\linewidth}{l*{8}{>{\centering\arraybackslash}X}}
            \toprule
            \textbf{Method} & \multicolumn{4}{c}{Replica} & \multicolumn{4}{c}{TAT} \\
            \cmidrule(lr){2-5} \cmidrule(lr){6-9}
            & \multicolumn{2}{c}{Rot ($^\circ$)} & \multicolumn{2}{c}{Trans (m)} & \multicolumn{2}{c}{Rot ($^\circ$)} & \multicolumn{2}{c}{Trans (m)} \\
            \cmidrule(lr){2-3} \cmidrule(lr){4-5} \cmidrule(lr){6-7} \cmidrule(lr){8-9}
            & Mean$\downarrow$ & Max$\downarrow$ & Mean$\downarrow$ & Max$\downarrow$ & Mean$\downarrow$ & Max$\downarrow$ & Mean$\downarrow$ & Max$\downarrow$ \\
            \midrule
            AnySplat       & 7.320 & 12.77 & 0.106 & 0.470 & 3.516 & 4.871 & 0.135 & 0.501 \\
            YoNoSplat      & 4.769 & 6.172 & 0.063 & 0.174 & 5.318 & 6.593 & 0.115 & 0.378 \\
            DepthAnything3 & 0.324 & 0.549 & \underline{0.004} & \underline{0.011} & \underline{0.682} & \underline{1.009} & \underline{0.030} & \underline{0.084} \\
            MapAnything    & 4.438 & 5.544 & 0.056 & 0.106 & 3.232 & 3.649 & 0.065 & 0.117 \\
            Ours (w/o TTO) & \underline{0.301} & \underline{0.494} & 0.006 & 0.014 & 1.636 & 2.040 & 0.047 & 0.145 \\
            Ours           & \textbf{0.220} & \textbf{0.316} & \textbf{0.002} & \textbf{0.006} & \textbf{0.262} & \textbf{0.356} & \textbf{0.020} & \textbf{0.057} \\
            \bottomrule
        \end{tabularx}%
        }
    \end{minipage}\hfill
    \begin{minipage}[t]{0.32\textwidth}
        \centering
        \scriptsize
        \captionof{table}{Chamfer distances for surface alignment on DTU.}
        \label{tab:surface_alignment}
        \renewcommand{\arraystretch}{1.42}
        \setlength{\tabcolsep}{1pt}
        \scalebox{\surfaceTableScale}{%
        \begin{tabularx}{0.85\linewidth}{>{\raggedright\arraybackslash}p{0.32\linewidth}*{3}{>{\centering\arraybackslash}X}}
            \toprule
            \textbf{Method} & \textbf{Acc.$\downarrow$} & \textbf{Com.$\downarrow$} & \textbf{Over.$\downarrow$} \\
            \midrule
            VoxelTTO\newline(wo. TTO) & 5.556 & 9.159 & 7.357 \\
            VoxelTTO\newline(Camera Enc.) & 3.719 & 4.853 & 4.286 \\
            VoxelTTO & \textbf{2.952} & \textbf{3.880} & \textbf{3.416} \\
            \bottomrule
        \end{tabularx}%
        }
    \end{minipage}
    \vspace{-4mm}
\end{table}


We use absolute translation error and rotation error to evaluate the poses. Translation error is the Euclidean distance between predicted and ground truth. Rotation error is the geodesic distance on the rotation manifold between prediction and ground truth: $e_R=\arccos\left((\operatorname{tr}\left(\mathbf{R}_{\mathrm{pred}}^{\top}\mathbf{R}_{\mathrm{gt}}\right)-1)/2\right)$.
As shown in \autoref{tab:pose_quantitative}, VoxelTTO freezes both the VFM backbone and pose decoder. Therefore, removing TTO yields pose-estimation accuracy comparable to DA3. The DA3 result in \autoref{tab:pose_quantitative} uses encoded camera parameters and is consequently slightly improved. 
Adding TTO substantially improves alignment between predicted and ground-truth poses.

\subsection{Surface Alignment}

As coupled errors in VFM poses and depth maps can displace the predicted point cloud from the ground-truth surface, further validation is carried on DTU. 
We compare VoxelTTO without TTO, VoxelTTO with DA3 camera-parameter encoding \cite{depthanything3} instead of TTO, and the full VoxelTTO model.
Following DTU\cite{dtu}, we report accuracy (Acc.), completeness (Com.), and their mean (Over.), which are all computed with Chamfer distance.
As shown in \autoref{tab:surface_alignment}, camera-parameter encoding suppresses misalignment to some extent. VoxelTTO further reduces accuracy, completeness, and overall distance.


\subsection{Ablation Studies}
\begin{figure*}[!h]
    \centering
    \begin{minipage}[t]{0.49\textwidth}
        \centering
        \includegraphics[width=0.99\linewidth]{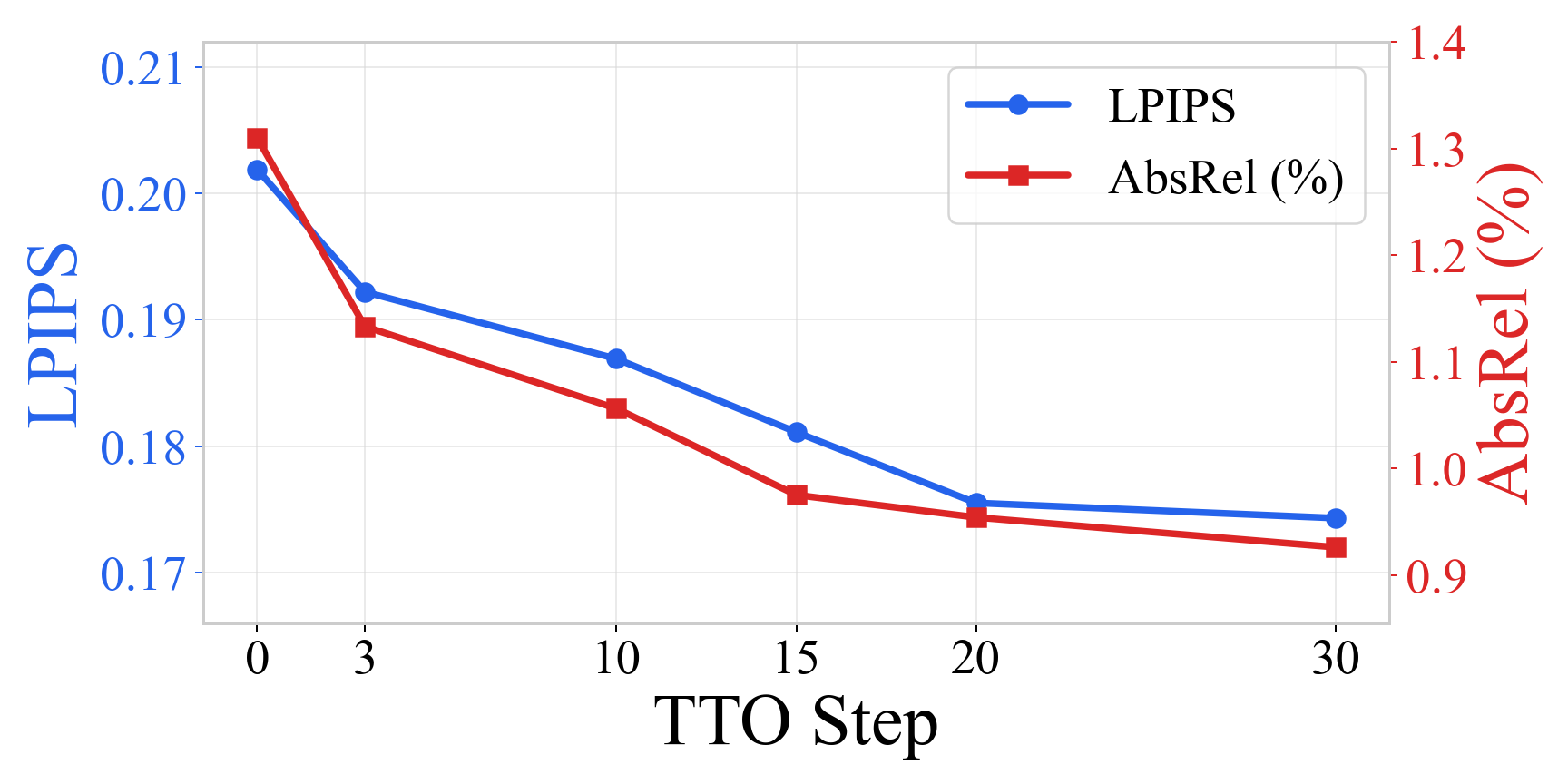}
        \captionof{figure}{Test-view RGB-D synthesis performance at different TTO iteration counts.}
        \label{fig:tco_step}
    \end{minipage}\hfill
    \begin{minipage}[t]{0.49\textwidth}
        \centering
        \includegraphics[width=0.99\linewidth]{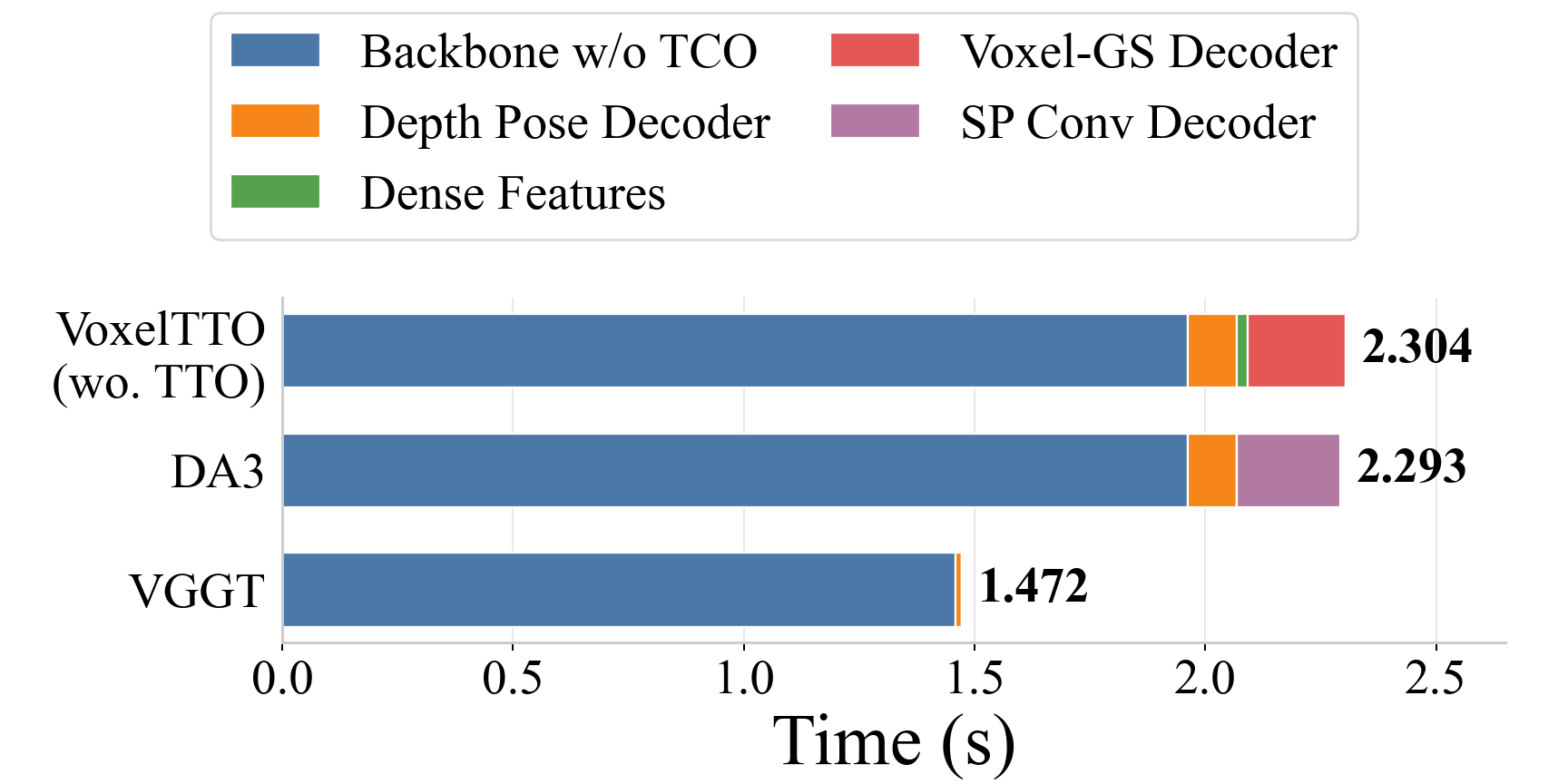}
        \captionof{figure}{Runtime of individual modules in VoxelTTO without TTO and in the baselines.}
        \label{fig:time_notto}
    \end{minipage}
    \vspace{-5mm}
\end{figure*}

\begin{table*}[!h]
    \centering
    \caption{Ablation studies on test-view RGB-D NVS under Replica dataset.}
    \label{tab:ablation_studies}
    \renewcommand{\arraystretch}{0.85}
    \setlength{\tabcolsep}{1pt}
    \newcommand{\ablationTableScale}{1.3}
    {\scriptsize\renewcommand{\tabularxcolumn}[1]{m{#1}}%
    \makebox[\textwidth][c]{\scalebox{\ablationTableScale}{%
    \begin{tabularx}{0.75\textwidth}{>{\centering\arraybackslash}m{0.14\textwidth}>{\centering\arraybackslash}m{0.10\textwidth}*{8}{>{\centering\arraybackslash}X}}
        \toprule
        \textbf{Experiment} & \textbf{Setting} & \textbf{PSNR}$\uparrow$ & \textbf{SSIM}$\uparrow$ & \textbf{LPIPS}$\downarrow$ & \shortstack{\textbf{RMSE}\\\textbf{(cm)}$\downarrow$} & \shortstack{\textbf{AbsRel}\\\textbf{(\%)}$\downarrow$} & \shortstack{$\boldsymbol{\delta}_{1.05}$$\uparrow$} & \shortstack{$\boldsymbol{\delta}_{1.10}$$\uparrow$} & \shortstack{$\boldsymbol{\delta}_{1.25}$$\uparrow$} \\
        \midrule
        \multirow{3}{*}{\shortstack[c]{Number of\\Gaussian Splits}} & 1 & 27.82 & 0.866 & \textbf{0.172} & 8.09 & 0.94 & 98.71 & 99.10 & 99.47 \\
        & 2 (Ours) & \textbf{28.54} & \underline{0.876} & \underline{0.172} & \textbf{6.95} & \underline{0.87} & \underline{98.87} & \underline{99.24} & \textbf{99.59} \\
        & 4 & \underline{28.35} & \textbf{0.876} & 0.179 & \underline{7.21} & \textbf{0.86} & \textbf{98.93} & \textbf{99.24} & \underline{99.54} \\
        \midrule
        \multirow{4}{*}{\shortstack[c]{Stochastic Solid\\Volume Rendering}} & $\mathbb{V}$-$\mathbb{V}$ (Ours) & \textbf{28.54} & \textbf{0.876} & \textbf{0.172} & \textbf{6.95} & \textbf{0.87} & \textbf{98.87} & \textbf{99.24} & \textbf{99.59} \\
        & $\mathbb{V}$-$\mathbb{R}$ & \underline{27.11} & \underline{0.865} & \underline{0.188} & \underline{12.01} & \underline{1.24} & \underline{97.96} & \underline{98.62} & \underline{99.15} \\
        & $\mathbb{R}$-$\mathbb{R}$ & 21.52 & 0.774 & 0.269 & 27.58 & 4.05 & 87.76 & 95.11 & 97.10 \\
        & $\mathbb{R}$-$\mathbb{V}$ & 22.37 & 0.798 & 0.260 & 15.82 & 3.04 & 89.32 & 97.03 & 98.66 \\
        \midrule
        \multirow{4}{*}{\shortstack[c]{Voxel\\Size}} & 0.0015 & \textbf{29.34} & \textbf{0.891} & \textbf{0.142} & \textbf{6.62} & \textbf{0.83} & \textbf{99.04} & \textbf{99.36} & \textbf{99.64} \\
        & 0.002 (Ours) & \underline{28.54} & \underline{0.876} & \underline{0.172} & \underline{6.95} & \underline{0.87} & \underline{98.87} & \underline{99.24} & \underline{99.59} \\
        & 0.003 & 26.78 & 0.856 & 0.211 & 9.45 & 1.18 & 98.32 & 98.88 & 99.32 \\
        & 0.005 & 23.44 & 0.798 & 0.290 & 15.22 & 2.45 & 93.17 & 97.15 & 98.47 \\
        \midrule
        \multirow{2}{*}{\shortstack[c]{Depth\\Loss}} & Ours & \textbf{28.54} & \textbf{0.876} & \underline{0.172} & \textbf{6.95} & \textbf{0.87} & \textbf{98.87} & \textbf{99.24} & \textbf{99.59} \\
        & No $\mathcal{L}_\text{depth}$ & \underline{28.25} & \underline{0.873} & \textbf{0.171} & \underline{7.49} & \underline{0.94} & \underline{98.67} & \underline{99.10} & \underline{99.53} \\
        \bottomrule
    \end{tabularx}
    }}%
    }
\end{table*}
All the ablation studies are shown in \autoref{tab:ablation_studies}.
For the self-splitting Gaussian decoder, we evaluate the \textbf{number of Gaussians splits} per voxel. Increasing the number of splits improves geometry and appearance, but the improvement from 2 to 4 splits is no longer substantial. 
To evaluate the effect of \textbf{stochastic solid volume rendering} on geometry, we compare it with 3DGS rasterization during both training and inference. Using 3DGS rasterization substantially reduces appearance and geometric accuracy under both stages.
Here, $\mathbb{V}$-$\mathbb{R}$ denotes stochastic solid volume-rendering supervision ($\mathbb{V}$) during training and 3DGS rasterization ($\mathbb{R}$) for NVS at inference.
We compare different voxel sizes in voxel-aligned 3DGS. Both appearance and geometric performance decline as \textbf{voxel size} increases. Balancing performance and computation, and following prior work \cite{jiang2025anysplat}, we set the size to 0.002m.
%
An RGB loss $\mathcal{L}_\text{rgb}$ is required to regress color. We therefore ablate the \textbf{depth loss} separately. Removing the depth loss slightly reduces geometric accuracy.
We further evaluate the effect of \textbf{TTO iterations}. As shown in \autoref{fig:tco_step}, test-view appearance performance (LPIPS) and geometric performance (AbsRel) improve substantially as the number of TTO iterations increases. We select 20 iterations by balancing accuracy and efficiency.

\subsection{Runtime}
\label{sec:time}
For ten $518\times518$ images, VoxelTTO without TTO takes 2.304\,s, compared with 2.293\,s for DA3, indicating negligible overhead from voxel decoding (\autoref{fig:time_notto}). We retain frozen attention computations in BF16 and LoRA computations in Float32, while baselines use their official precision. BF16 reduces backbone runtime from 1.959\,s to 0.1956\,s. Each TTO step requires 0.2662\,s forward and 0.075\,s backward, so 20 steps complete in under 7\,s and total runtime is under 7.4\,s.

\section{Conclusion}
We presented VoxelTTO, a voxel-aligned feed-forward 3DGS framework with test-time optimization. First, it reduces the excessive overlap and artifacts of pixel-aligned feed-forward 3DGS by aggregating dense pixel features into global voxel features. A sparse 3D voxel U-Net and self-splitting decoder regress Gaussian attributes. Stochastic solid volume rendering replaces 3DGS rasterization during training and inference, improving the geometric accuracy of the Gaussian scene.
Second, pose-supervised TTO adapts lightweight LoRA modules in the VFM backbone to align predicted camera parameters with the provided camera priors. 
Extensive experiments on DTU, Replica, and T\&T show that VoxelTTO reconstructs geometrically accurate 3DGS scenes and achieves SOTA test-view RGB and depth rendering accuracy.

\subsection*{AI use statement}


In this work, we did not use generative AI tools for any tasks requiring disclosure under the ICLR 2027 AI Policy for Authors. We used generative AI tools solely for language editing and improving the readability of the manuscript; create or edit software code. We have reviewed all AI-assisted work. All AI-assisted edits were carefully reviewed by the authors to ensure correctness and to preserve the intended technical meaning. We take responsibility for the final content of this work, including text, claims or artifacts produced with the aid of generative AI.



\subsection*{Reproducibility statement}

The train dataset can be found in:

ScanNet\cite{dai2017scannet}\url{https://scannetpp.mlsg.cit.tum.de/scannetpp}; 
Infinigen\cite{infinigen2023infinite}\url{https://github.com/princeton-vl/infinigen}; 
ARKitScenes\cite{baruch2021arkitscenes}\url{https://github.com/apple-aiml-research/ARKitScenes}; 
DL3DV\cite{ling2024dl3dv}\url{https://dl3dv-10k.github.io/DL3DV-10K/}

The eval dataset can be found in 

Replica\cite{replica19arxiv}\url{https://github.com/facebookresearch/replica-dataset};
Tanks and Temples\cite{tatdataset}\url{https://www.tanksandtemples.org/};
DTU\cite{dtu}\url{http://roboimagedata.compute.dtu.dk/?page_id=36}

The code can be found in

\url{https://anonymous.4open.science/r/VoxelTTO-0432} and \url{https://anonymous.4open.science/r/Geometry-Grounded-Gaussian-Splatting-1B46}

\subsubsection*{Acknowledgments}
Acknowledgment
The authors gratefully acknowledge the financial support from the Zhejiang Provincial “Pioneer” and “Leading Goose + X” Science and Technology Program (Grant No. 2026C02A1220), the Shanghai Science and Technology Action Plan (Grant No. 21JM0010300), the Shanghai Aerospace Science and Technology Innovation Fund (SAST) (Grant No. 2021-037), the Special Fund for Technology Innovation Support Projects of Shanghai (Grant Nos. 2021-cyxt-kj1, XTCX-KJ-2022-37, and HCXBCY-2023-046), and the National Defense Basic Scientific Research Program of China (Grant No. JCKY2021606B002).

\bibliography{iclr2027_conference}
\bibliographystyle{iclr2027_conference}


\end{document}

%% file: math_commands.tex
\usepackage{amsmath,amsfonts,bm}

\def\eqref#1{equation~\ref{#1}}

\def\1{\bm{1}}

\DeclareMathAlphabet{\mathsfit}{\encodingdefault}{\sfdefault}{m}{sl}
\SetMathAlphabet{\mathsfit}{bold}{\encodingdefault}{\sfdefault}{bx}{n}

